\documentclass{article}

\usepackage{microtype}
\usepackage{graphicx}
\usepackage{subcaption}
\usepackage{booktabs} 
\usepackage{enumitem}
\usepackage{bm}             
\usepackage{soul}           

\usepackage{hyperref}

\usepackage[preprint]{icml2026}

\usepackage{amsmath}
\usepackage{amssymb}
\usepackage{mathtools}
\usepackage{amsthm}
\usepackage{ulem}

\usepackage[capitalize,noabbrev]{cleveref}

\theoremstyle{plain}

\theoremstyle{definition}

\theoremstyle{remark}

\usepackage[textsize=tiny]{todonotes}

\icmltitlerunning{\texttt{GHOST}: Unmasking Phantom States in \texttt{Mamba2}}

\begin{document}

\twocolumn[
\icmltitle{\texttt{GHOST}: Unmasking Phantom States in \texttt{Mamba2} \\ via Grouped Hidden-state Output-aware Selection \& Truncation}



\icmlsetsymbol{equal}{*}

\begin{icmlauthorlist}
\icmlauthor{Michael Menezes}{rice}
\icmlauthor{Anastasios Kyrillidis}{rice}
\end{icmlauthorlist}

\icmlaffiliation{rice}{Department of Computer Science, Rice University, Houston, TX, United States}

\icmlcorrespondingauthor{Anastasios Kyrillidis}{anastasios@rice.edu}

\icmlkeywords{Machine Learning, ICML}

\vskip 0.3in
]



\printAffiliationsAndNotice{}  


\begin{abstract}
While \texttt{Mamba2}'s expanded state dimension enhances temporal modeling, it incurs substantial inference overhead that saturates bandwidth during autoregressive generation. Standard pruning methods fail to address this bottleneck: unstructured sparsity leaves activations dense, magnitude-based selection ignores runtime dynamics, and gradient-based methods impose prohibitive costs. We introduce \texttt{GHOST} (Grouped Hidden-state Output-aware Selection and Truncation), a structured pruning framework that approximates control-theoretic balanced truncation using only forward-pass statistics. By jointly measuring controllability and observability, \texttt{GHOST} rivals the fidelity of gradient-based methods without requiring backpropagation. 
As a highlight, on models ranging from 130M to 2.7B parameters, our approach achieves a 50\% state-dimension reduction with approximately 1 perplexity point increase on WikiText-2. Code is available at \url{https://anonymous.4open.science/r/mamba2_ghost-7BCB/}.
\end{abstract}

\section{Introduction}
\label{sec:introduction}

Driven by scaling laws, the transition from \texttt{Mamba1} to \texttt{Mamba2} increased the SSM state dimension $N$ from 16 to 128 \citep{dao_state_2024}; for a 1.3B model, this swells the recurrent state from $\approx 12$ MB to $\approx 100$ MB. While enhancing temporal modeling, this $8 \times$ expansion creates an inference bottleneck by saturating memory bandwidth and degrading cache locality \citep{asif_perfmamba_2025}. Effective post-training compression is essential to democratize the deployment of large-scale SSMs. However, pruning the state dimension of \texttt{Mamba2} presents challenges that expose the limitations of transformer-based compression paradigms. 

First, \textit{unstructured methods can be insufficient for inference acceleration.} While second-order optimizers like \texttt{OBC} \citep{frantar2023optimalbraincompressionframework} and \texttt{SparseGPT} \citep{frantar_SparseGPT_2023}, or simple activation-aware metrics like \texttt{Wanda} \citep{sun_simple_2024}, effectively sparsify projection weights, their original formulations generally yield \textit{dense activations}. Since the product of a sparse weight row and a dense input vector is non-zero, the resulting projected input remains dense. Consequently, \texttt{Mamba2}'s recurrent state $\bm{H}_t$ (to be defined later in text) remains fully populated, foregoing reductions in memory bandwidth consumption.

\begin{figure}[!t]
\centering
\includegraphics[width=\columnwidth]{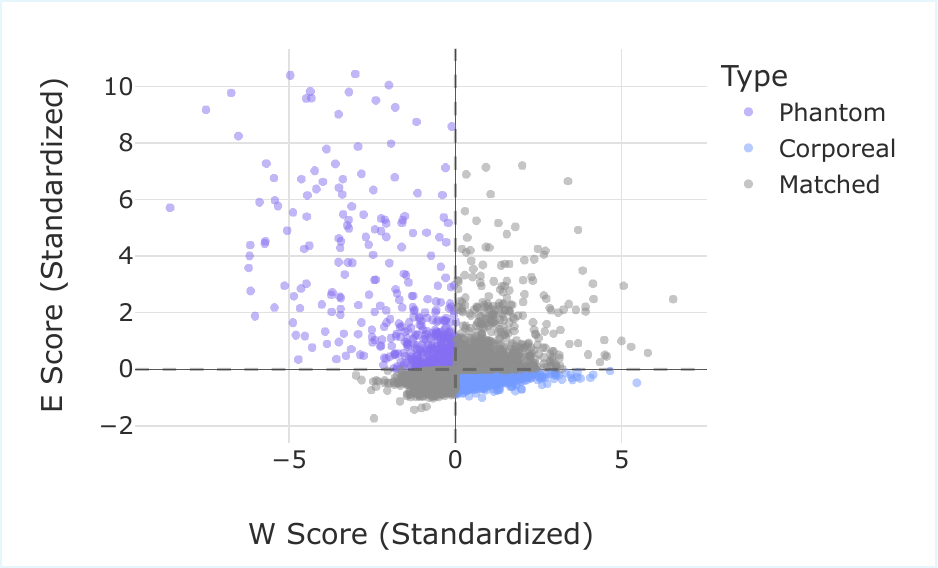}
\caption{\textbf{The Proxy Failure: Static Magnitude vs. Dynamic Energy.} We analyze the correlation between weight-based importance and actual runtime usage for \texttt{Mamba2}-1.3B. Each point represents one of the 6,144 hidden states ($48$ layers with $128$ states each). The $x$-axis depicts the standardized static score derived from projection weights ($W_{\text{score}} = \sqrt{\| (\bm{W}_{\bm{B}})_{g,i} \|_2 \| (\bm{W}_{\bm{C}})_{g,i} \|_2}$ for state $i$ in group $g$), while the $y$-axis tracks the standardized dynamic energy as in Eq. \eqref{eq:salience}. The lack of positive correlation reveals two critical failure modes: \textbf{Phantom States} (top-left), which are highly active despite low weight norms, and \textbf{Corporeal States} (bottom-right), which have large weights but low utilization. \texttt{GHOST} is designed to identify the former and prune the latter.}
\label{fig:weight_energy_correlation}
\end{figure}

Second, \textit{static magnitude pruning is often blind to system dynamics.} Standard magnitude pruning \citep{saikumar2025signalcollapseoneshotpruning} evaluates a state channel based on fixed weight norms rather than its dynamic energy. This reliance on static proxies creates a ``blind spot'' where the pruner cannot distinguish between a state's theoretical capacity and its actual utilization in \texttt{Mamba}. We categorize the resulting failure modes into \textit{corporeal states}, false-negative channels which appear structurally significant but are inert under dynamics, and \textit{phantom states}, false-positive channels which exhibit low weight norms but high activity. As illustrated in \Cref{fig:weight_energy_correlation}, there is often no positive correlation between a state's static magnitude and its dynamic energy. So, standard pruning inadvertently discards high-activity phantom states while preserving inert corporeal ones.

Third, \textit{gradient-based solutions can be expensive.} While structured methods like \texttt{Taylor} pruning \citep{ghattas_pruning_2025} offer high fidelity by estimating loss sensitivity, 
they require computing and storing gradients for the full computational graph. This consumes 45 GB VRAM for a 1.3B model that exceeds e.g., the standard 40 GB A100 capacity. In contrast, our proposed method utilizes only $15$ GB. To offset prohibitive costs, gradient-based methods necessitate ``one-shot'' estimation, which suffers from distribution shift as upstream pruning invalidates downstream gradients \citep{frantar2023optimalbraincompressionframework}.

\textbf{This paper.}
We introduce Grouped Hidden-state Output-aware Selection and Truncation, or \texttt{GHOST}, a structured pruning framework that approximates balanced truncation \citep{moore_principal_1981}, using only forward-pass statistics. By computing the product of empirical controllability (how effectively inputs drive a channel) and observability (how strongly a channel influences outputs) from calibration data, \texttt{GHOST} identifies and preserves ``living'' states that are simultaneously reachable and impactful.

\texttt{GHOST} operates within \texttt{Mamba2}'s Grouped Query Attention (GQA) structure, where groups of heads share dynamics parameters. We pool saliency scores across groups within each layer, enabling adaptive capacity allocation: groups modeling complex dynamics retain more states, while redundant groups are aggressively pruned. Sequential layer-by-layer processing with activation updates mitigates distribution shift, and the entire procedure requires only two forward passes per layer. No gradients. No Hessians. No quadratic memory overhead.

Our main contributions are as follow:
\begin{itemize}[leftmargin=*]
\item We propose \texttt{GHOST}, a data-driven structured pruning framework for \texttt{Mamba2} that utilizes inter-group thresholding to prune the SSM state dimension $N$ based on measured controllability and observability.
\item We demonstrate that \texttt{GHOST} solves the ``proxy failure'' of \texttt{Magnitude} pruning, successfully exorcising \textit{corporeal} and preserving \textit{phantom} states.
\item We provide empirical evidence that \texttt{GHOST} outperforms \texttt{Magnitude}, and \texttt{Random} baselines, and rivals the performance of \texttt{Taylor} pruning, while requiring \textit{zero gradient computation} and a fraction of the cost.
\end{itemize}

\section{Preliminaries}
\label{sec:preliminaries}
Here, we provide a brief overview of the \texttt{Mamba2} architecture. A more detailed exposition is found in \Cref{app:preliminaries}. Readers familiar with these topics may proceed to \Cref{sec:related_work}.

Given input $\bm{u}_t \in \mathbb{R}^M$, the model applies RMSNorm and a linear transformation to produce the gate $\bm{z}_t$, input $\bm{x}_t$, dynamics parameters $\bm{B}_t, \bm{C}_t$, and timescale $\bm{\Delta}_t$\footnote{We follow the notation used in the literature, but we acknowledge that there is ambiguity about what is a vector (usually boldface, lowercase letters) and we have to make exceptions like $\bm{\Delta}_t \in \mathbb{R}^H$. A complete notation table is provided in Appendix \ref{app:table_notation}.}:
\begin{equation}
[\bm{z}_t; \bm{x}_t; \bm{B}_t; \bm{C}_t; \bm{\Delta}_t] = \bm{W}_{\text{in}}\operatorname{RMSNorm}(\bm{u}_t) + \bm{b}_{\text{in}}, \label{eq:mamba2_projection}
\end{equation}
where $\bm{z}_t, \bm{x}_t \in \mathbb{R}^{H \cdot P}, \bm{\Delta}_t \in \mathbb{R}^{H}$ are unique to each of the $H$ heads, while $\bm{B}_t, \bm{C}_t \in \mathbb{R}^{G \times N}$ are shared within $G$ groups of $K = H / G$ heads. Following a depthwise convolution and SiLU activation on the signal and dynamics branches (yielding $\bm{x}'_t, \bm{B}'_t, \bm{C}'_t$), the sequence undergoes discretized recurrence using learned parameters $\bm{A}_h = \bm{a}_h \bm{I} \in \mathbb{R}^N$ and $\bm{D} \in \mathbb{R}^{H}$. For head $h$ belonging to group $g$, the hidden state $\bm{H}_{t,h} \in \mathbb{R}^{P \times N}$ and output $\bm{y}^{\text{SSM}}_{t,h} \in \mathbb{R}^P$ are updated as:
\begin{align}
\bm{H}_{t,h} &= \overline{\bm{A}}_{t, h} \odot \bm{H}_{t-1,h} + \bm{x}'_{t,h} \otimes (\overline{\bm{B}}_{t, h})^\top, \label{eq:hidden_update}\\
\bm{y}^{\text{SSM}}_{t,h} &= \bm{H}_{t,h} (\bm{C}'_{t,g})^\top + \bm{D}_h \bm{x}'_{t,h}, \label{eq:output_compute}
\end{align}
where $\overline{\bm{A}}_{t, h} = \exp(\bm{\Delta}_{t,h} \log (\bm{A}_h))$ and $\overline{\bm{B}}_{t, h} = \bm{\Delta}_{t,h} \bm{B}'_{t,g}$. The final block output is obtained by gating $\bm{y}^{\text{SSM}}_t$ with $\operatorname{SiLU} (\bm{z}_t)$, normalizing, projecting with $\bm{W}_{\text{out}} \in \mathbb{R}^{M \times H \cdot P}$, and adding the residual connection.

\section{Related Work}
\label{sec:related_work}

Existing SSM compression methodologies are often incompatible with \texttt{Mamba2}'s architectural innovations or orthogonal to state-dimension pruning. For instance, \texttt{SparseSSM} \citep{tuo_sparsessm_2025} targets the diagonal $\bm{A}$ matrix of \texttt{Mamba1} to determine state importance; this metric is unavailable in \texttt{Mamba2}, as its scalar-identity formulation lacks state-specific granularity. Similarly, timescale-based methods like \texttt{PerfMamba} \citep{asif_perfmamba_2025} and $\Delta$-guided pruning \citep{anonymous2025dtp} operate at the granularity of heads; while effective for head-pruning, they cannot resolve intra-head state redundancy. Coarser techniques like \texttt{Mamba-Shedder} \citep{munoz_mamba-shedder_2025} remove entire architectural blocks, lacking the granularity required to compress the state dimension $N$. Finally, while gradient-based structured pruning discussed in \Cref{sec:introduction} is theoretically applicable, it suffers from the \textit{masked distribution shift} problem \citep{frantar2023optimalbraincompressionframework} and prohibitive memory overheads.

\section{Methodology}
\label{sec:method}

\begin{figure*}[t]
\centering
\includegraphics[width=\textwidth]{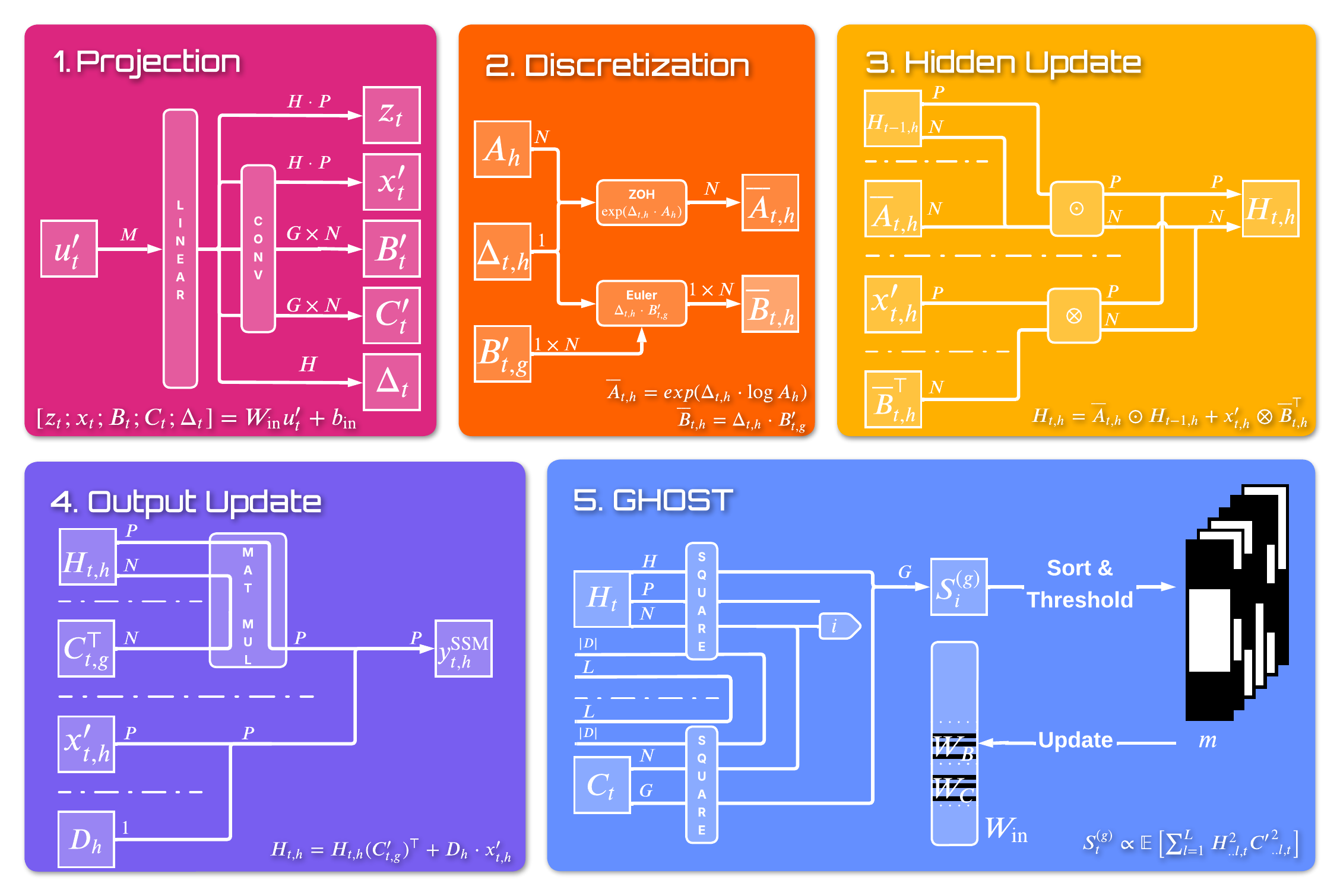} \vspace{-0.5cm}
\caption{Overview of the \texttt{Mamba2} forward pass with \texttt{GHOST}. The input $u'_t$ is projected into intermediate variables (Subfigure 1.) and discretized parameters (Subfigure 2.) to update the hidden state $\bm{H}_{t,h}$ (Subfigure 3.) and compute the final SSM output $\bm{y}^{\text{SSM}}_{t,h}$ (Subfigure 4.). Concurrently, the \texttt{GHOST} mechanism computes scores $\bm{S}_t^{(g)}$ and applies sorting and thresholding to generate a binary mask $\bm{m} \in \mathbb{R}^{G \cdot N}$ that induces sparsity in the initial projection (Subfigure 5.).} \vspace{-0.3cm}
\label{fig:ghost}
\end{figure*}

We propose \texttt{GHOST} (Grouped Hidden-state Output-aware Selection and Truncation), a structured pruning framework designed to compress the state dimension $N$ of \texttt{Mamba2} as \Cref{fig:ghost} illustrates.

\subsection{Problem Formulation}
\label{subsec:problem_formulation}
We consider a pre-trained \texttt{Mamba2} model $f(\cdot; \bm{\theta})$, where we abstractly denote trainable parameters as $\bm{\theta}$. Our objective is to identify a binary mask $\bm{M} \in \{0, 1\}^{N_{\text{layers}} \times G \cdot N}$ that zeroes out specific channels in the state dimension $N$. Pruning $N$ directly reduces the size of the recurrent state $\bm{H}_t$, yielding strictly lower activation memory and FLOPs during recurrence. We minimize prediction error on a calibration set $\mathcal{D}_{\text{cal}}$, subject to a target state sparsity $\kappa \in [0,1]$:
\begin{equation*}
\begin{aligned}
\min_{\bm{M}} \quad & \mathbb{E}_{x \sim \mathcal{D}_{\text{cal}}} [\mathcal{L}(f(x; \bm{\theta}|_{\bm{M}}))]\\
\textrm{s.t.} \quad & \frac{\|\bm{M}\|_0}{N_{\text{layers}} \cdot G \cdot N} \leq 1 - \kappa.
\end{aligned}
\end{equation*}
Pruning the state dimension directly reduces the memory footprint of $\bm{H}_t$ from $(H, P, N)$ to $(H, P, N')$ where $N' = (1-\kappa)N$, with proportional reductions in memory bandwidth during autoregressive generation.

\subsection{Theoretical Motivation: Balanced Truncation}
\label{subsec:balanced_truncation}
\texttt{GHOST} approximates the principles of \textit{balanced truncation} \citep{moore_principal_1981} to identify and prune redundant state channels. Fundamentally, this approach seeks to retain states that serve as effective conduits for information flow, characterized by: (1) \textbf{Controllability:} How much does each previous input impact the current state; and (2) \textbf{Observability:} How much does the current state impact future outputs. In the context of standard state-space models, the most critical states are those that are both well used under the input history and have a significant impact on future predictions.

The projection in \Cref{eq:mamba2_projection} characterizes \texttt{Mamba2} as a Linear Time-\textit{Varying} (LTV) system governed by the input-dependent terms $\bm{\Delta}_t, \bm{B}'_t$, and $\bm{C}'_t$. Moreover, the SiLU activation that follows the depthwise convolution renders the system non-commutative. These properties preclude the use of standard algebraic solutions (e.g., Lyapunov equations), necessitating a time-varying, data-driven approach.

\subsection{Empirical Estimation via Covariance and Hessians}
\label{subsec:gramians}
To adapt controllability and observability for the \texttt{Mamba2} architecture, we use instantaneous empirical Gramians \citep{lall2002subspace, a11070091} evaluated at time $t$ over a sequence of length $L$.

\textbf{Hidden State Covariance as Controllability.} We quantify the extent to which the input history uses a state by analyzing the statistics of the hidden states themselves. Since $\bm{H}_{t,h}$ is the accumulation of processed inputs, its magnitude represents the energy transferred from the input sequence. Per \citet{a11070091}, we therefore define the empirical measure of controllability as the \textit{hidden state covariance}, which captures the variance of the state induced by the data.

\textbf{Output Energy Hessian as Observability.} To quantify the influence of the current state on future outputs, we examine the curvature of the output energy with respect to the state. We define the observability metric as the \textit{Hessian of $\sum_{k = t}^{L} \frac{1}{2} ( \bm{y}^{\text{SSM}}_{k,h,p} )^2$ with respect to the hidden state $\bm{H}_{t,h,p}$}. For the immediate output $\bm{y}^{\text{SSM}}_{t,h,p} = \bm{H}_{t,h,p} (\bm{C}'_{t,g})^\top$, the energy is given by $\mathcal{E} = \frac{1}{2} ( \bm{y}^{\text{SSM}}_{t,h,p} )^2$. The Hessian with respect to the state is:
\begin{equation*}
\nabla^2_{\bm{H}_{t,h,p}} \mathcal{E} = (\bm{C}'_{t,g})^\top \bm{C}'_{t,g}.
\end{equation*}
While a full observability metric would sum these Hessians over all future time steps $k \geq t$, we adopt a local approximation based on the immediate projection $(\bm{C}'_{t,g})^\top \bm{C}'_{t,g}$. This is justified by the decay inherent in the discretized transition dynamics, which suggests that the contribution of future outputs to the current state's observability diminishes exponentially \citep{gwak2025layeradaptivestatepruningdeep}. This instantaneous Hessian captures the dominant term of the full observability Gramian while providing a causal, computationally efficient proxy that eliminates the need for a backward pass over the temporal horizon.

\textbf{Saliency Scoring.} Given that $\bm{A}_h$ is scalar-identity per head, the state channel recurrences are independent and we restrict our analysis to the diagonal terms of controllability and observability. For a specific state channel $i \in [N]$ within layer $j$, group $g$, head $k$, and head channel $p$, we define the instantaneous diagonal scores as:
\begin{align*}
\bm{P}^{(j, g)}_{k, p, i} (t) = \left( \bm{H}^{(j)}_{t, h_k, p, i} \right)^2, \qquad \bm{Q}^{(j, g)}_{i} (t) = \left( \bm{C}'^{(j)}_{t, g, i} \right)^2.
\end{align*}
Moreover, we define the instantaneous saliency $\bm{S}^{(j, g)}_{k, p, i} (t) = \bm{P}^{(j, g)}_{k, p, i} (t) \cdot \bm{Q}^{(j, g)}_{i} (t)$ as the product of these terms, mirroring the construction of Hankel singular values in balanced truncation \citep{moore_principal_1981}. To obtain a robust global estimate, we aggregate this score over the sequence length $L$ and the calibration dataset $\mathcal{D}_{\text{cal}}$. Reducing the GQA structure via expectation, the final saliency score for state channel $i$ is:
\begin{equation}
\label{eq:salience}
\bm{S}^{(j, g)}_i = \sqrt{\frac{1}{Z} \sum_{d=1}^{|\mathcal{D}_{\text{cal}}|} \sum_{t=1}^{L} \sum_{k=1}^{K} \sum_{p=1}^{P} S^{(j, g)}_{k, p, i} (t | d)},
\end{equation}
where $Z = |\mathcal{D}_{\text{cal}}| \cdot K \cdot P$ is the normalization constant. The final square root aligns our metric with the theoretical definition of Hankel singular values ($\sigma_i \approx \sqrt{\lambda_i(\bm{P}\bm{Q})}$). Beyond maintaining theoretical fidelity, this transformation restores the dimensionality of the score to the linear scale of the input features, facilitating direct magnitude comparisons.

\subsection{Impact on Local Loss}
\label{subsec:local_loss}
Although we derived our salience score using control theory, it is consistent with a loss preservation perspective. Due to the GQA structure of \texttt{Mamba2}, pruning a state channel $i$ within group $g$ only affects the subset of $K$ heads associated with that group. Let $\mathcal{L}^{(g)}_t$ be the local reconstruction error at time $t$ for group $g$, defined as the squared Frobenius norm of the difference between the pruned and original outputs for the affected heads. Following standard analysis in SSM literature~\cite{gu2024mambalineartimesequencemodeling, dao_transformers_2024}, we set $\bm{D} = 0$ as the feedthrough term does not contribute to state dynamics, the primary targets of our pruning.
\begin{align*}
\mathcal{L}^{(g)}_t (\bm{H}'_t) &= \sum_{k=1}^K \| \bm{y}'^{\text{SSM}}_{t,h_k} - \hat{\bm{y}}^{\text{SSM}}_{t,h_k} \|_F^2 \\
&= \sum_{k=1}^K \| \bm{H}'_{t,h_k} (\bm{C}'_{t,g})^\top - \bm{H}_{t,h_k} (\bm{C}'_{t,g})^\top \|_F^2,
\end{align*}
where $h_k$ denotes the global index of the $k$-th head in group $g$. The error simplifies to the impact of the state perturbation $\delta \bm{H}_{t,h_k} = \bm{H}'_{t,h_k} - \bm{H}_{t,h_k}$. Pruning channel $i$ corresponds to zeroing out the $i$-th column of the hidden state matrix for these specific heads. Substituting $\delta \bm{H}_{t,h_k,\cdot,i} = -\bm{H}_{t,h_k,\cdot,i}$ yields:
\begin{align*}
\mathcal{L}^{(g)}_t &= \sum_{k=1}^K \| \delta \bm{H}_{t,h_k} (\bm{C}'_{t,g})^\top \|_F^2 \\
&= \sum_{k=1}^K \sum_{p=1}^P (\bm{H}_{t,h_k,p,i})^2 (\bm{C}'_{t,g,i})^2.
\end{align*}
We seek a pruning decision robust across the data distribution. Averaging over calibration set $\mathcal{D}_{\text{cal}}$ and summing over sequence length $L$, we define the \textit{expected cumulative error}:
\begin{align*}
\mathbb{E} [\mathcal{L}] &= \mathbb{E} \left[ \sum_{t=1}^L \mathcal{L}^{(g)}_t \right] \\
&= \frac{1}{| \mathcal{D}_{\text{cal}} |} \sum_{d=1}^{|\mathcal{D}_{\text{cal}}|} \sum_{t=1}^{L} \sum_{k=1}^{K} \sum_{p=1}^{P} (\bm{H}_{t|d,h_k,p,i})^2 (\bm{C}'_{t|d,g,i})^2.
\end{align*}
Observing that $\mathbb{E} [ \mathcal{L} ] \propto \bm{S}_i^2$, we conclude that minimizing the \texttt{GHOST} saliency score $\bm{S}_i$ is equivalent to minimizing the expected increase in local mean-squared error.

\subsection{Inter-Group Thresholding and Pruning}
\label{subsec:selection}
\texttt{GHOST} performs \textit{inter-group thresholding}. We pool the scores from all $G$ groups within the current layer:
\begin{equation*}
\mathcal{S}_{pool}^{(j)} = \bigcup_{g=1}^{G} \{ \bm{S}^{(j, g)}_{i} \mid i \in [N] \}.
\end{equation*}
We sort these $G \times N$ scores and determine a global threshold $\tau_j$ based on the target sparsity for that layer. This allows the model to dynamically allocate state capacity; groups modeling complex dynamics may retain more channels $r > N(1-\kappa)$, while redundant groups are heavily pruned.

To implement this, we employ a soft pruning strategy where projection weights $\bm{W}_{\{B, C\}}$ and convolution filters are zeroed out according to the binary mask derived from $\tau_j$. The model is then updated sequentially, using the modified activations from layer $j$ to calibrate layer $j+1$, ensuring subsequent layers adapt to the sparsified dynamics.

\subsection{Complexity Analysis}
\label{subsec:complexity}

\texttt{GHOST}, \Cref{alg:ghost}, is computationally efficient, requiring only two forward passes over the calibration data. In practice, we compute raw sums rather than the root mean square shown in Eq. \eqref{eq:salience}, as the raw values are rank equivalent and faster to compute.

\begin{algorithm}[h]
\caption{\texttt{GHOST}: Given \texttt{Mamba2} layer \texttt{mixer} and inputs $\mathcal{X}_{\text{in}}$, we prune state dimension $N$ to $(1 - \kappa) \cdot N$.}
\label{alg:ghost}
\begin{algorithmic}   
\STATE $\mathcal{S}_{\text{pool}} \leftarrow \emptyset$
\STATE $\bm{S}^{(g)} \leftarrow \bm{0}_{N} \ \forall g$

\FOR{$\bm{u} \in \mathcal{X}_{\text{in}}$}
\FOR{$t=1$ {\bfseries to} $L$}

\STATE $\_, \bm{H}_t, \bm{C}_t \leftarrow \texttt{mixer}(\bm{u}_t)$ // transient states
\STATE $\bm{S}^{(g)} \leftarrow \bm{S}^{(g)} + \sum_{k, p} (\bm{H}_{t,h_k,p})^2 \cdot (\bm{C}_{t,g})^2 \ \forall g$

\ENDFOR
\ENDFOR

\STATE $\mathcal{S}_{\text{pool}} \leftarrow \{ \bm{S}^{(g)} : g \in [G]\}$ // inter-group pool

\STATE $\bm{m} \leftarrow$ mask of $\kappa \cdot N$ states with lowest $\mathcal{S}_{\text{pool}}$ rank
\STATE $\bm{W}_{\{B, C\}}, \bm{C}_{\{B, C\}} \leftarrow \left( \bm{W}_{\{B, C\}}, \bm{C}_{\{B, C\}} \right) \odot \bm{m}$

\STATE $\mathcal{X}_{\text{out}}, \_, \_ \leftarrow \texttt{mixer}(\mathcal{X}_{\text{in}})$ // update activations
\STATE \textbf{return} $\texttt{mixer}, \mathcal{X}_{out}$
\end{algorithmic}
\end{algorithm}

\textbf{Time Complexity:} $O(|\mathcal{D}_{\text{cal}}| \cdot L \cdot G \cdot K \cdot P \cdot N)$ per layer. This matches the cost of a standard inference pass, as we accumulate squared values during computation.

\textbf{Space Complexity:} $O(G \cdot N)$. We only store $N$ statistics per group, avoiding the $O(N^2)$ storage required by full-covariance methods or Hessian-based pruning.

\section{Experiments}






We evaluate \texttt{GHOST} across multiple dimensions: sparsity levels, sequence lengths, model scales, downstream tasks, and out-of-distribution robustness. Unless otherwise noted, we fixed random seeds to 42 and used \texttt{Mamba2}-1.3B with Huggingface default precision (float32) and sequence length (2048) \citep{wolf2020huggingfacestransformersstateoftheartnatural}. We calibrated on 128 data samples of WikiText-2 \citep{merity2016pointersentinelmixturemodels} at a batch size of 1 to save memory and compressed to 50\% sparsity. We measured model performance using EleutherAI's LM Evaluation Harness \citep{eval-harness}. All experiments were run on a single H100 80 GB GPU.

\begin{figure*}[h]
\centering
\includegraphics[width=\textwidth]{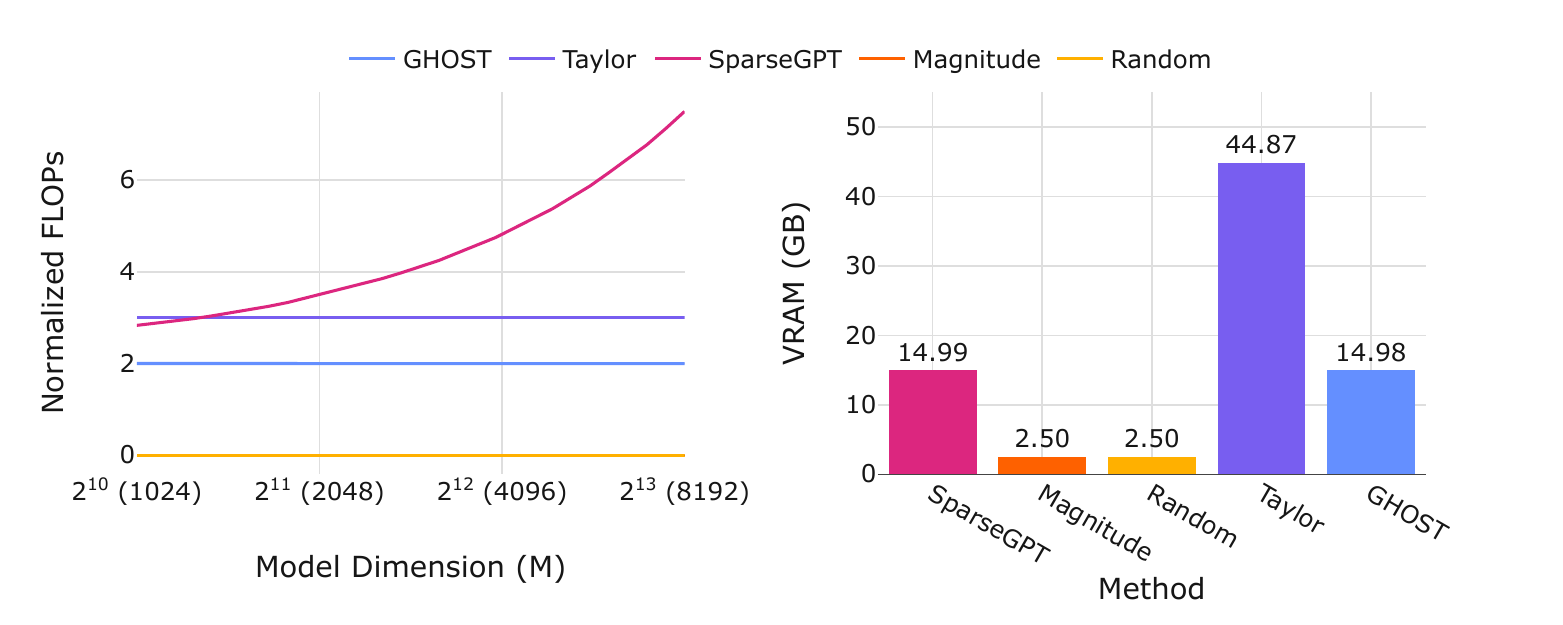}
\caption{\textbf{Time and Space Efficiency.} (Left) FLOPs normalized by the computation required for a single forward pass. (Right) Peak VRAM requirements by method. Note that \texttt{Taylor} exceeds the memory capacity of 40 GB cards.}
\label{fig:data_efficiency}
\end{figure*}

\Cref{fig:data_efficiency} illustrates the computational trade-offs of the evaluated methods. Regarding time efficiency (left), \texttt{Magnitude} and \texttt{Random} are essentially free. Among the data-driven methods, while all share linear complexity with respect to the pruning target $N$, the critical differentiator is their scaling with the dimension $M$. Here, \texttt{SparseGPT} scales quadratically, whereas \texttt{Taylor} and \texttt{GHOST} remain linear. Regarding memory constraints (right), \texttt{Taylor}'s reliance on gradient computation incurs significant overhead, exceeding the capacity of standard A100 40 GB GPUs.

\subsection{Impact of Sparsity Levels}
\label{subsec:sparsity_sweep}

To assess the robustness of \texttt{GHOST} against varying degrees of compression, we conducted a sparsity sweep ranging from 10\% to 90\% sparsity. We compared the post-pruning perplexity on WikiText-2 against competing methods. The dense baseline attained a perplexity of $13.17$. 

The results presented in \Cref{tab:sparsity_sweep} demonstrate a distinct trend: while \texttt{Taylor} exhibited robust performance at lower sparsity levels, it destabilized as the pruning threshold increased. To mitigate the high computational cost of backpropagation, \texttt{Taylor} relies on one-shot score computation. However, this efficiency trade-off renders it susceptible to catastrophic collapse, as it fails to account for the distribution shifts induced by pruned layers. In contrast, \texttt{GHOST}'s lower cost allows for sequential updates to ease this effect. Moreover, by prioritizing energy preservation in the feature space, \texttt{GHOST} remained competitive with \texttt{Taylor} and \texttt{SparseGPT} in low-sparsity regimes. Crucially, despite operating under computational and structural constraints, \texttt{GHOST} maintained a smooth degradation curve, yielding a viable model even at 70\% sparsity: a regime where other structural baselines suffered from perplexity explosions.

\begin{table}[!t]
\centering
\caption{\textbf{Sparsity Sweep.} Perplexity comparison on WikiText-2 across increasing sparsity levels. While top methods perform comparably at moderate sparsity (50\%), \texttt{GHOST} demonstrates significantly better retention of model capabilities in the high-sparsity regime ($>70\%$), avoiding the catastrophic breakdown observed in \texttt{Taylor}.}
\label{tab:sparsity_sweep}
\begin{tabular}{lccccc}
\toprule
& \multicolumn{5}{c}{\textbf{Target Sparsity}} \\
\cmidrule(lr){2-6}
\textbf{Method} & \textbf{10\%} & \textbf{30\%} & \textbf{50\%} & \textbf{70\%} & \textbf{90\%} \\
\midrule
\texttt{SparseGPT} & 13.17 & 13.19 & 13.25 & 13.52 & 15.51 \\
\midrule
Magnitude & 427.7 & 22.39 & 29.34 & 39.96 & 69.49 \\
Random & 13.52 & 15.04 & 17.77 & 30.14 & 64.21 \\
\texttt{Taylor} & \textbf{13.18} & \textbf{13.26} & \textbf{13.94} & 4255 & 4690 \\
\texttt{GHOST} & 13.24 & 13.51 & 14.23 & \textbf{16.16} & \textbf{25.07} \\
\bottomrule
\end{tabular}
\end{table}

\subsection{Sequence Length Generalization}
\label{subsec:length_gen}

For computing empirical Gramians, \texttt{GHOST} assumes a target sequence length. We investigated how this impacts sequence length generalization by calibrating all methods on short contexts, $L_{\text{cal}} = 128$, and evaluated perplexity on increasingly long contexts up to $L_{\text{eval}}=2048$. 

\Cref{tab:length_generalization} shows that \texttt{GHOST} generalizes well: perplexity decreases with longer evaluation contexts, following the same trend as the dense model. In contrast, \texttt{Taylor}'s perplexity increases with sequence length, reaching 1613 at $L_{\text{eval}} = 2048$---a dramatic failure suggesting that gradient-based scores computed on short contexts do not transfer to long-range dependencies.


\begin{table}[!ht]
\centering
\caption{\textbf{Length Generalization.} Comparison of perplexity degradation when evaluating on sequence lengths longer than the calibration context. We denote calibration free methods with *.}
\label{tab:length_generalization}
\begin{tabular}{l|c|cccc}
\toprule
\textbf{Method} & \textbf{$\bm{L_{\text{cal}}}$} & \multicolumn{4}{c}{\textbf{Evaluation Length ($\bm{L_{\text{eval}}}$)}} \\
& \textbf{128} & \textbf{256} & \textbf{512} & \textbf{1024} & \textbf{2048} \\
\midrule
Dense* & 23.04 & 18.52 & 15.88 & 14.20 & 13.18 \\
\texttt{SparseGPT} & 23.11 & 18.59 & 15.96 & 14.29 & 13.27 \\
\midrule
Magnitude* & 32.76 & 27.96 & 26.38 & 27.32 & 29.34 \\
Random* & 27.81 & 22.89 & 20.17 & 18.59 & 17.77 \\
\texttt{Taylor} & 317.3 & 386.9 & 497.8 & 797.0 & 1613 \\
\texttt{GHOST} & \textbf{23.99} & \textbf{19.46} & \textbf{16.85} & \textbf{15.22} & \textbf{14.27} \\
\bottomrule
\end{tabular}
\end{table}

\subsection{Sequence Length Robustness}
\label{subsec:length_robust}

\begin{figure}[h]
\centering
\includegraphics[width=\columnwidth]{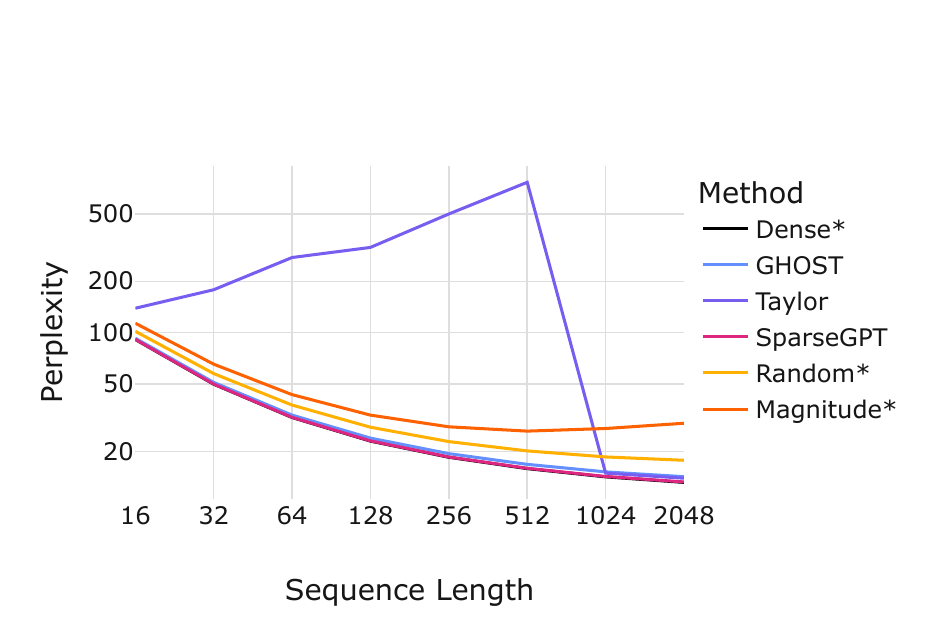}
\caption{\textbf{Length Robustness.} Comparison of perplexity degradation when evaluating on increasingly shorter contexts. We denote calibration free methods with *.}
\label{fig:length_robust}
\end{figure}

Beyond sequence length generalization, we evaluated robustness when calibrating and testing on identical, progressively shorter sequence lengths. As illustrated in the log-log plot in \Cref{fig:length_robust}, while all methods exhibit a gentle increase in perplexity at lower sequence lengths, \texttt{Taylor} displays a distinct anomalous behavior. Specifically, we observe a sharp phase transition where \texttt{Taylor} diverges significantly in the $L \in [16, 512]$ regime, underperforming even trivial baselines like \texttt{Magnitude} and \texttt{Random}. Outside this instability window, \texttt{Taylor} rapidly converges back to the performance profile of top methods. In contrast, \texttt{GHOST} maintains consistent stability across the sweep, closely tracking the trends of \texttt{SparseGPT} and \texttt{Dense} baselines. See \Cref{app:plot_tabulations} \Cref{tab:length_robust} for a tabulation.

\subsection{Scaling Laws}
\label{subsec:scaling}
We evaluated the efficacy of pruning across varying model capacities, ranging from 130M to 2.7B parameters. Smaller models typically exhibited lower parameter redundancy, making them significantly more sensitive to the removal of state dynamics. As shown in \Cref{tab:scaling}, this regime exposes critical instabilities in \texttt{Taylor} which failed catastrophically on the 130M and 370M models, resulting in perplexity explosions ($>10^{4}$), likely due to the noisy gradient signals inherent in smaller state-space models.

In contrast, \texttt{GHOST} demonstrated superior stability. It outperformed all baselines on the 130M and 370M models, proving that its feature-based selection criterion effectively identifies essential dynamics even when redundancy is scarce. While \texttt{Taylor} recovered performance at larger scales (1.3B and 2.7B) where gradient signals presumably stabilize, \texttt{GHOST} remained the only method to provide consistent, usable post-pruning performance across the entire scaling spectrum.

\begin{table}[!t]
\centering
\caption{\textbf{Scaling Laws.} Perplexity comparison across \texttt{Mamba2} model sizes. While \texttt{Taylor} collapses on smaller architectures due to gradient instability, \texttt{GHOST} remains robust across all scales.}
\label{tab:scaling}
\begin{tabular}{cccccc}
\toprule
{\textbf{Method}} & \textbf{130M}      & \textbf{370M}     & \textbf{780M}     & \textbf{1.3B}     & \textbf{2.7B}     \\
\midrule
Dense                                  & 25.85 & 18.13 & 14.97 & 13.17 & 11.46 \\
\texttt{SparseGPT}                              & 26.17 & 18.34 & 15.08 & 13.25 & 11.51 \\
\midrule
Magnitude                              & 910.0 & 120.3  & 30.57 & 29.34 & 20.04 \\
Random                                 & 45.60 & 23.71 & 5197 & 17.77 & 15.93 \\
\texttt{Taylor}                                 & 1E16 & 12647  & \textbf{15.13} & \textbf{13.94} & \textbf{11.50} \\
\texttt{GHOST}                         & \textbf{29.06}  & \textbf{19.96} & 16.13 & 14.23 & 12.14 \\
\bottomrule
\end{tabular}
\end{table}

\subsection{Zero-Shot Performance}
\label{subsec:zero_shot}

\begin{figure}[h]
\centering
\includegraphics[width=\columnwidth]{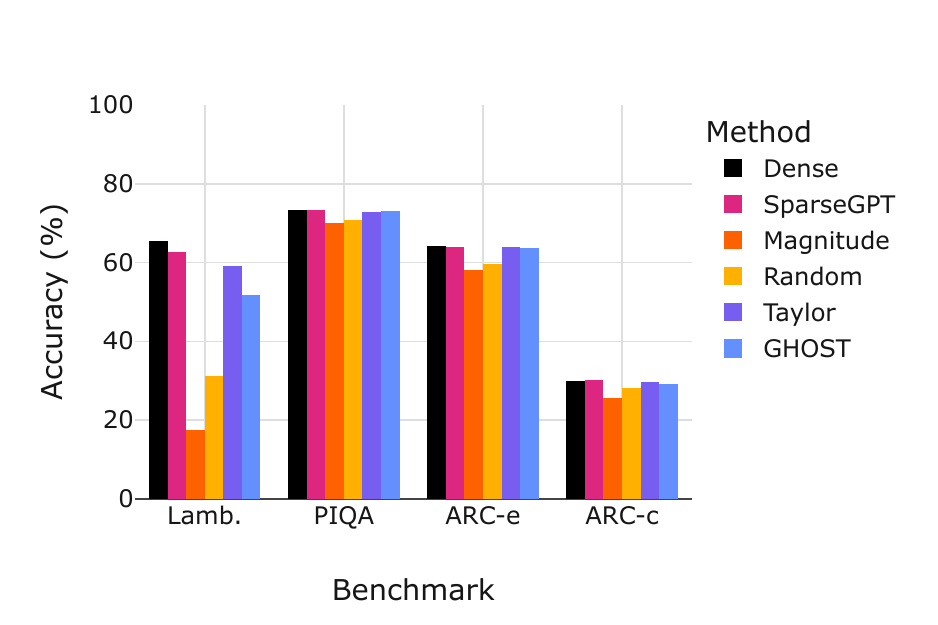}
\caption{\textbf{Zero-Shot Performance.} Accuracy on Lambada, PIQA, ARC-e, and ARC-c.}
\label{fig:zero_shot}
\end{figure}

To analyze flexibility across data domains, we calibrated and evaluated all methods on various zero shot tasks: Lambada \citep{paperno2016lambadadatasetwordprediction}, PIQA \citep{bisk2019piqareasoningphysicalcommonsense}, ARC-e, and ARC-c \citep{clark2018thinksolvedquestionanswering}. See \Cref{fig:zero_shot}.

As expected, the unpruned \texttt{Dense} model provides an upper bound for performance, while \texttt{SparseGPT} generally retains high fidelity. Among the structured pruning baselines, naive approaches such as \texttt{Magnitude} and \texttt{Random} pruning suffer catastrophic performance drops on the Lambada dataset, falling to 17.50\% and 31.21\% respectively. \texttt{Taylor} pruning demonstrates robustness, achieving the highest accuracy among compressed methods on Lambada (59.23\%), ARC-e (63.85\%), and ARC-c (29.69\%). Our proposed method, \texttt{GHOST}, remains highly competitive, even outperforming \texttt{Taylor} on the PIQA dataset with an accuracy of 73.12\%, which is within 0.22\% of the dense baseline. We use a bar chart to highlight the close trend between \texttt{Taylor} and \texttt{GHOST}. See \Cref{app:plot_tabulations} \Cref{tab:zero_shot} for a tabulation.

\subsection{Out-of-Distribution (OOD) Robustness}
\label{subsec:ood_robustness}

Tensor-wise, Hessian-based pruning often overfits the curvature of the specific calibration domain (e.g., English text) \citep{tang2025gaprunegradientalignmentpruningdomainaware}. \texttt{GHOST}, acting as a structural regularizer based on system dynamics, should offer better generalization to unseen distributions. We calibrated on WikiText-2 and evaluated zero-shot performance on code generation (HumanEval) and multiple choice math problems (MMLU Elementary Mathematics). As hypothesized, \texttt{GHOST} outperforms the baselines on OOD tasks (\Cref{tab:ood_robustness}). This suggests that \texttt{GHOST} identifies channels that are fundamentally important to the State Space Model's operation, rather than those that are merely important for predicting the next word in English prose.

\begin{table}[h]
\centering
\caption{\textbf{OOD Robustness.} Performance on OOD datasets, HumanEval and MMLU Elementary Math, after calibrating solely on WikiText-2. \texttt{GHOST} shows the smallest gap between in-domain (ppl) and out-of-domain (acc) performance.}
\label{tab:ood_robustness}
\begin{tabular}{lccc}
\toprule
& \textbf{In-Domain} & \multicolumn{2}{c}{\textbf{Out-of-Domain}} \\
\cmidrule(lr){2-2} \cmidrule(lr){3-4}
\textbf{Method} & \textbf{Text} & \textbf{Code} & \textbf{Math} \\
\midrule
Dense & 13.17 & 6.71 & 21.43 \\
\texttt{SparseGPT} & 13.25 & 5.49 & 22.75 \\
\midrule
Magnitude & 29.34 & 0.61 & \textbf{25.66} \\
Random & 17.77 & 4.27 & 21.43 \\
\texttt{Taylor} & \textbf{13.94} & 4.88 & 22.49 \\
\texttt{GHOST} & 14.23 & \textbf{5.49} & \textbf{25.66} \\
\bottomrule
\end{tabular}
\end{table}

\subsection{Calibration Data Efficiency}
\label{subsec:data_efficiency}

We evaluated the perplexity of the pruned models while varying the number of calibration samples $|\mathcal{D}_{\text{cal}}| \in \{8, 16, 32, 64, 128\}$. As shown in \Cref{tab:data_efficiency}, all methods show equivalent sample efficiency.

\begin{table}[h]
\centering
\caption{\textbf{Robustness to Data Scarcity.} Perplexity on WikiText-2 with varying calibration set sizes.}
\label{tab:data_efficiency}
\begin{tabular}{lccccc}
\toprule
& \multicolumn{5}{c}{\textbf{Calibration Samples ($\bm{|\mathcal{D}_{\text{cal}}|}$)}} \\
\cmidrule(lr){2-6}
\textbf{Method} & \textbf{8} & \textbf{16} & \textbf{32} & \textbf{64} & \textbf{128} \\
\midrule
\texttt{SparseGPT} & 13.27 & 13.26 & 13.25 & 13.25 & 13.25 \\
\midrule
\texttt{Taylor}  & 13.30 & 13.33 & 13.30 & 13.35 & 13.94 \\
\texttt{GHOST} & 14.28 & 14.26 & 14.23 & 14.23 & 14.23 \\
\bottomrule
\end{tabular}
\end{table}

\subsection{Saliency Ablation}
\label{subsec:saliency_ablation}

To evaluate the importance of the mixed controllability and observability score, we conducted a sparsity sweep ranging from 10\% to 90\% sparsity while changing the definition of $\bm{S}^{(j, g)}_i (t)$ between $\bm{P}^{(j, g)}_{k, p, i} (t)$, $\bm{Q}^{(j, g)}_{i} (t)$, and $\bm{P}^{(j, g)}_{k, p, i} (t) \cdot \bm{Q}^{(j, g)}_{i} (t)$. As shown in \Cref{tab:saliency_ablation}, the joint measure (\texttt{GHOST}) consistently scores best and is most stable.

\begin{table}[h]
\centering
\caption{\textbf{Saliency Ablation.} Perplexity comparison of \texttt{GHOST} on WikiText-2 across sparsity levels while changing the definition of $\bm{S}^{(j, g)}_i (t)$ between $\bm{P}^{(j, g)}_{k, p, i} (t)$, $\bm{Q}^{(j, g)}_{i} (t)$, and $\bm{P}^{(j, g)}_{k, p, i} (t) \cdot \bm{Q}^{(j, g)}_{i} (t)$.}
\label{tab:saliency_ablation}
\begin{tabular}{lccccc}
\toprule
& \multicolumn{5}{c}{\textbf{Target Sparsity}} \\
\cmidrule(lr){2-6}
\textbf{Method} & \textbf{10\%} & \textbf{30\%} & \textbf{50\%} & \textbf{70\%} & \textbf{90\%} \\
\midrule
$\bm{P}^{(j, g)}_{k, p, i} (t)$ & 13.26 & 13.64 & 14.56 & 16.71 & 2158 \\
$\bm{Q}^{(j, g)}_{i} (t)$ & 13.37 & 14.23 & 16.16 & 19.83 & 33.12 \\
\texttt{GHOST} & \textbf{13.24} & \textbf{13.51} & \textbf{14.23} & \textbf{16.16} & \textbf{25.07} \\
\bottomrule
\end{tabular}
\end{table}


\section{Discussion and Limitations}
\label{sec:discussion}

\textbf{When Does \texttt{GHOST} Fail?}
\texttt{GHOST} relies on the assumption that controllability and observability can be estimated from forward-pass statistics. This assumption weakens in several regimes.
At extreme sparsity ($>90\%$), perplexity degrades to 25, suggesting the retained states cannot capture essential dynamics; as shown in \Cref{app:extra_plots} \Cref{fig:energy_elbow}'s elbow plot, this sparsity level forces the removal of highly utilized states. Similarly, small models (130M) suffer a sharper perplexity increase (29.06 vs.\ 25.85 dense) than larger ones, likely due to lower representational redundancy.

\textbf{Computational Overhead.}
While \texttt{GHOST} avoids gradient computation, it requires extracting hidden states during the forward pass, adding modest overhead. On \texttt{Mamba2}-1.3B without custom kernels, \texttt{GHOST} calibration takes 8 minutes (2 forward passes over 128 samples), compared to $<$1 minute for \texttt{Magnitude}/\texttt{Random} and 8 minutes for \texttt{Taylor} (which additionally requires $>$45\,GB VRAM).

\textbf{Generalization to Other Architectures.}
\texttt{GHOST} is designed for \texttt{Mamba2}'s specific structure: scalar-identity $\bm{A}$ and grouped dynamics. Extending to other SSM variants (S4, H3, \texttt{Mamba1}) may require adapting the Gramian approximations. The core principle of jointly measuring input-to-state and state-to-output energy should transfer, but the specific instantiation depends on architectural details.

\textbf{Physical Speedup.}
This paper demonstrates that \texttt{GHOST} identifies prunable states while preserving model quality. Converting soft pruning to actual speedup requires either (1) sparse kernel support, or (2) physically removing pruned channels and adjusting downstream layers. We leave these optimizations to future work and delegate to \cite{asif_perfmamba_2025} for empirical evidence of state reduction's speedup.

\section{Conclusion}

In this work, we addressed the critical inference bottlenecks of large-scale \texttt{Mamba2} models by targeting the recurrent state dimension $N$ for structured compression. We identified limitations of existing pruning paradigms, specifically the ``blind spots'' of static magnitude metrics and the prohibitive costs of gradient-based analysis. By bridging deep learning with control theory, we proposed \texttt{GHOST}, a novel pruning framework that approximates balanced truncation to select state channels based on their dynamic energy transfer.

Our method effectively distinguishes between truly salient dynamics from misleading weight magnitude assessments, i.e. corporeal and phantom states. Extensive experiments confirm that \texttt{GHOST} achieves state-of-the-art trade-offs between sparsity and accuracy, rivaling the performance of expensive second-order methods without the need for gradient computation or massive GPU memory. \texttt{GHOST} not only democratizes the deployment of \texttt{Mamba2} by reducing memory bandwidth consumption but also demonstrates remarkable robustness to sequence length variations and distribution shifts. Future work may extend these control-theoretic principles to other architectures, further refining the intersection of system identification and neural network compression or seek system level optimizations.




\section*{Impact Statement}

This paper presents work whose goal is to advance the field of Machine
Learning. There are many potential societal consequences of our work, none
which we feel must be specifically highlighted here.

\bibliography{references}
\bibliographystyle{icml2026}

\newpage
\appendix
\onecolumn
\crefalias{section}{appendix}

\section{Preliminaries}
\label{app:preliminaries}
We provide expanded background on state space models and the \texttt{Mamba} architecture.

\textbf{State Space Models.}
State space models describe dynamical systems via a hidden state that evolves over time. In continuous time, a linear time-invariant (LTI) system takes the form:
\begin{align}
\dot{\bm{h}}(t) &= \bm{A}\bm{h}(t) + \bm{B}x(t), \label{eq:app_continuous_state} \\
y(t) &= \bm{C}\bm{h}(t) + D x(t), \label{eq:app_continuous_output}
\end{align}
where $\bm{h}(t) \in \mathbb{R}^N$ is the hidden state, $x(t) \in \mathbb{R}$ is the input, $y(t) \in \mathbb{R}$ is the output, $\bm{A} \in \mathbb{R}^{N \times N}$ governs state dynamics, $\bm{B} \in \mathbb{R}^{N \times 1}$ maps inputs to states, $\bm{C} \in \mathbb{R}^{1 \times N}$ maps states to outputs, and $D \in \mathbb{R}$ is a direct feedthrough term.

For sequences, time is discretized using step size $\Delta \in \mathbb{R}$ converting \Cref{eq:app_continuous_state,eq:app_continuous_output} into
\begin{align}
\bm{h}_t &= \overline{\bm{A}}\bm{h}_{t-1} + \overline{\bm{B}}x_t, \label{eq:app_discrete_state} \\
y_t &= \bm{C}\bm{h}_t + D x_t, \label{eq:app_discrete_output}
\end{align}
where $\overline{\bm{A}} = \exp(\Delta \bm{A})$ represents the Zero-Order Hold (ZOH) discretization, and $\overline{\bm{B}} \approx \Delta \bm{B}$ is the Euler discretization often used for efficiency.

The recurrence in \Cref{eq:app_discrete_state,eq:app_discrete_output} can be viewed as a linear RNN, or equivalently unrolled into a global convolution for parallel training. This dual view underlies the efficiency of SSM-based architectures: recurrent for inference, convolutional for training.

\textbf{The \texttt{Mamba} Architecture.}
\texttt{Mamba}~\citep{gu2024mambalineartimesequencemodeling} extends classical SSMs with selective state spaces, where the dynamics parameters $\bm{B}_t$, $\bm{C}_t$, and discretization step are input-dependent rather than fixed. This selectivity enables content-based reasoning: the model can modulate how strongly each input affects the hidden state and how the state influences outputs.

\textit{\texttt{Mamba2} and GQA Semantics.}
\texttt{Mamba2}~\citep{dao_state_2024} restructures the computation to leverage tensor cores, achieving significant speedups over \texttt{Mamba1}. 
A key architectural change is the adoption of Grouped Query Attention (GQA) semantics: the $H$ attention heads are partitioned into $G$ groups, with each group of $K = H/G$ heads sharing the same dynamics parameters $\bm{B}_t$ and $\bm{C}_t$. This sharing reduces parameter count and enables more efficient computation, but creates structural constraints for pruning. Now, removing a state channel affects all heads within a group.

\textit{\texttt{Mamba2} Block Structure.}
Let $M$ be the model dimension. So, the expanded dimension $R = E \cdot M$ where $E$ is usually 2, and $R = H \cdot P$ where $H$ is the number of heads and $P$ is the head dimension. Let $N$ be the state dimension. Given layer input $\bm{u}_t \in \mathbb{R}^M$, a \texttt{Mamba2} block proceeds as follows.

Using pre-layer normalization, $\bm{u}'_t = \operatorname{RMSNorm} (\bm{u}_t)$ where for an input vector $\bm{v} \in \mathbb{R}^V$, the Root Mean Square Normalization is defined as:
\begin{equation*}
\text{RMSNorm}(\bm{v}) = \frac{\bm{v}}{\sqrt{\frac{1}{V} \sum_{i=1}^V \bm{v}_i^2 + \epsilon}} \odot \bm{\gamma},
\end{equation*}
with $\bm{\gamma} \in \mathbb{R}^V$ being a learnable scale parameter and $\epsilon > 0$ a small constant for numerical stability.

A single linear projection generates all required quantities:
\begin{equation}
[\bm{z}_t; \bm{x}_t; \bm{B}_t; \bm{C}_t; \bm{\Delta}_t] = \bm{W}_{\text{in}}\bm{u}'_t + \bm{b}_{\text{in}},
\label{eq:app_mamba2_projection}
\end{equation}
where $\bm{z}_t, \bm{x}_t \in \mathbb{R}^{H \cdot P}$ are the gate and input signals (unique per head), $\bm{B}_t, \bm{C}_t \in \mathbb{R}^{G \times N}$ are the dynamics matrices (shared per group), and $\bm{\Delta}_t \in \mathbb{R}^H$ contains the discretization steps (unique per head).\footnote{We follow the notation used in the literature, but we acknowledge that there is ambiguity about what is a vector (usually boldface, lowercase letters) and we have to make exceptions like $\bm{\Delta}_t \in \mathbb{R}^H$. A complete notation table is provided in Appendix \ref{app:table_notation}.}

The input and dynamics undergo depthwise convolution and SiLU activation, where $\operatorname{SiLU}(x) = x \cdot \sigma(x) = \frac{x}{1 + e^{-x}}$:
\begin{equation}
[\bm{x}'_t; \bm{B}'_t; \bm{C}'_t] = \operatorname{SiLU}(\operatorname{Conv1D}([\bm{x}_t; \bm{B}_t; \bm{C}_t])).
\label{eq:app_mamba2_conv}
\end{equation}

The discretized recurrence then computes, for each head $h$ in group $g$:
\begin{align}
\overline{\bm{A}}_{t,h} &= \exp(\bm{\Delta}_{t,h} \cdot \log(\bm{A}_h)), \label{eq:app_discretize_A} \\
\overline{\bm{B}}_{t,h} &= \bm{\Delta}_{t,h} \cdot \bm{B}'_{t,g}, \label{eq:app_discretize_B} \\
\bm{H}_{t,h} &= \overline{\bm{A}}_{t,h} \odot \bm{H}_{t-1,h} + \bm{x}'_{t,h} \otimes ( \overline{\bm{B}}_{t,h} )^\top, \label{eq:app_hidden_update} \\
\bm{y}^{\text{SSM}}_{t,h} &= \bm{H}_{t,h} (\bm{C}'_{t,g})^\top + \bm{D}_h \cdot \bm{x}'_{t,h}, \label{eq:app_output_compute}
\end{align}
where $\bm{A}_h = \bm{a}_h \cdot \bm{I} \in \mathbb{R}^N$ is the learned (negative, ensuring stability) diagonal of the continuous-time transition matrix for head $h$, $\bm{H}_{t,h} \in \mathbb{R}^{P \times N}$ is the hidden state, $\odot$ denotes elementwise multiplication (broadcast across $P$), $\otimes$ denotes outer product, and $\bm{D}_h$ is the feedthrough scalar. 

Finally, \texttt{Mamba2} gates $\bm{y}^{\text{gated}}_t = \operatorname{RMSNorm} ( \bm{y}^{\text{SSM}}_t \odot \text{SiLU}(\bm{z}_t) )$, projects $\bm{y}_t = \bm{W}_{\text{out}} \bm{y}^{\text{gated}}_t + \bm{b}_{\text{out}}$ and adds a residual connection to yield the layer output $\bm{u}_{t+1} = \bm{y}_t + \bm{u}_t$.

\textit{The Inference Bottleneck.}
The hidden state $\bm{H}_t$ has shape $(H, P, N)$ per token. 
For \texttt{Mamba2}-1.3B with $H=64$, $P=64$, $N=128$, this is $64 \times 64 \times 128 \times 4 = 2.1$ MB per layer in float32, or approximately 100 MB across 48 layers. During autoregressive generation, this state must be loaded and stored at every step, creating a memory-bandwidth bottleneck that structured pruning of $N$ directly addresses.

\section{Notation}
\label{app:table_notation}

\Cref{tab:notation} summarizes the mathematical notation used throughout this paper.

\begin{table}[h]
\centering
\caption{\textbf{Summary of Notation.}}
\label{tab:notation}
\begin{tabular}{cl}
\toprule
\textbf{Symbol} & \textbf{Description} \\
\midrule
\multicolumn{2}{l}{\textit{Model Dimensions}} \\
$M$ & Model dimension \\
$R$ & Expanded dimension ($R = E \cdot M = H \cdot P$) \\
$E$ & Expansion factor (typically 2) \\
$H$ & Number of attention heads \\
$P$ & Head dimension \\
$G$ & Number of groups \\
$K$ & Heads per group ($K = H/G$) \\
$N$ & State dimension \\
$L$ & Sequence length \\
\midrule
\multicolumn{2}{l}{\textit{Inputs and Outputs}} \\
$\bm{u}_t \in \mathbb{R}^M$ & Layer input at time $t$ \\
$\bm{x}_t, \bm{x}'_t \in \mathbb{R}^{H \cdot P}$ & Input signal (before/after conv and activation) \\
$\bm{z}_t \in \mathbb{R}^{H \cdot P}$ & Gate signal \\
$\bm{y}_t \in \mathbb{R}^M$ & Layer output \\
$\bm{y}^{\text{SSM}}_t \in \mathbb{R}^{H \cdot P}$ & SSM output (before gating) \\
\midrule
\multicolumn{2}{l}{\textit{SSM Dynamics}} \\
$\bm{A} \in \mathbb{R}^{N \times N}$ & Continuous-time state transition matrix \\
$\bm{B} \in \mathbb{R}^{N \times 1}$ & Input-to-state matrix \\
$\bm{C} \in \mathbb{R}^{1 \times N}$ & State-to-output matrix \\
$\bm{B}_t, \bm{B}'_t \in \mathbb{R}^{G \times N}$ & Input-dependent dynamics (before/after activation) \\
$\bm{C}_t, \bm{C}'_t \in \mathbb{R}^{G \times N}$ & Input-dependent dynamics (before/after activation) \\
$\bm{\Delta}_t \in \mathbb{R}^H$ & Discretization step sizes \\
$\overline{\bm{A}}, \overline{\bm{B}}$ & Discretized dynamics matrices \\
$\bm{H}_{t,h} \in \mathbb{R}^{P \times N}$ & Hidden state (per head) \\
$\bm{h}_t \in \mathbb{R}^N$ & Hidden state vector \\
$\bm{D}_h \in \mathbb{R}$ & Feedthrough scalar for head $h$ \\
\midrule
\multicolumn{2}{l}{\textit{Projections and Weights}} \\
$\bm{W}_{\text{in}} \in \mathbb{R}^{(2\cdot R + 2 \cdot G \cdot N + H) \times M}, \bm{b}_{\text{in}} \in \mathbb{R}^{2\cdot R + 2 \cdot G \cdot N + H}$ & Input projection weights and biases \\
$\bm{W}_{\text{out}} \in \mathbb{R}^{M \times H \cdot P}, \bm{b}_{\text{out}} \in \mathbb{R}^M$ & Output projection weights and biases \\
$\bm{W}_B \in \mathbb{R}^{G \cdot N \times M}, \bm{W}_C \in \mathbb{R}^{G \cdot N \times M}$ & Projection weights for $\bm{B}$ and $\bm{C}$ \\
\midrule
\multicolumn{2}{l}{\textit{Gramians and Saliency}} \\
$\bm{P} \in \mathbb{R}^{N \times N}$ & Controllability Gramian \\
$\bm{Q} \in \mathbb{R}^{N \times N}$ & Observability Gramian \\
$\bm{P}^{(j,g)}_{k,p,i}(t)$ & Controllability of state $i$ (layer $j$, group $g$, head $k$, chan $p$) \\
$\bm{Q}^{(j,g)}_i(t)$ & Observability of state $i$ (layer $j$, group $g$) \\
$\bm{S}^{(j,g)}_i$ & Saliency score for state $i$ (layer $j$, group $g$) \\
$\mathcal{S}_{\text{pool}}$ & Pooled saliency scores across groups \\
\midrule
\multicolumn{2}{l}{\textit{Pruning}} \\
$\bm{m}$ & Binary pruning mask \\
$\kappa$ & Target sparsity ratio \\
$\tau_j$ & Pruning threshold for layer $j$ \\
$\mathcal{D}_{\text{cal}}$ & Calibration dataset \\
\midrule
\multicolumn{2}{l}{\textit{Operations}} \\
$\odot$ & Element-wise (Hadamard) product \\
$\otimes$ & Outer product \\
$\|\cdot\|_F$ & Frobenius norm \\
$\|\cdot\|_0$ & $\ell_0$ pseudo-norm (number of nonzeros) \\
$\sigma(\cdot)$ & Sigmoid function \\
$\text{SiLU}(x)$ & $x \cdot \sigma(x) = \frac{x}{1+e^{-x}}$ \\
\bottomrule
\end{tabular}
\end{table}

\clearpage

\section{Tabulations of Plots.}
\label{app:plot_tabulations}

\begin{table}[h]
\centering
\caption{\textbf{Length Robustness.} Comparison of perplexity degradation when evaluating on increasingly shorter contexts. We denote calibration free methods with *.}
\label{tab:length_robust}
\begin{tabular}{l|cccccccc}
\toprule
\textbf{Method} & \multicolumn{8}{c}{\textbf{Calibration and Evaluation Length ($\bm{L}$)}} \\
& \textbf{16} & \textbf{32} & \textbf{64} & \textbf{128} & \textbf{256} & \textbf{512} & \textbf{1024} & \textbf{2048} \\
\midrule
Dense* & 91.24  & 49.81  & 31.73  & 23.04  & 18.52  & 15.88  & 14.20 & 13.17 \\
\texttt{SparseGPT} & 91.37  & 49.87  & 31.79  & 23.11  & 18.58  & 15.95  & 14.27 & 13.25 \\
\midrule
Magnitude*      & 113.7             & 65.29             & 43.22             & 32.76             & 27.96             & 26.38             & 27.32             & 29.34 \\
Random*         & 101.9             & 57.48             & 37.54             & 27.81             & 22.89             & 20.17             & 18.59             & 17.77 \\
\texttt{Taylor}          & 139.3             & 178.9             & 276.7             & 317.3             & 499.0             & 766.3             & \textbf{14.94}    & \textbf{13.94} \\
\texttt{GHOST}  & \textbf{92.75}    & \textbf{51.01}    & \textbf{32.76}    & \textbf{23.99}    & \textbf{19.44}    & \textbf{16.82}    & 15.19             & 14.23 \\
\bottomrule
\end{tabular}
\end{table}

\begin{table}[h]
\centering
\caption{\textbf{Zero-Shot Performance.} Accuracy on Lambada, PIQA, ARC-e, and ARC-c.}
\label{tab:zero_shot}
\begin{tabular}{c|cccc}
\toprule
\textbf{Method} & \textbf{Lamb.}   & \textbf{PIQA}    & \textbf{ARC-e}   & \textbf{ARC-c} \\
\midrule
Dense                                  & 65.55 & 73.34 & 64.23 & 29.95 \\
\texttt{SparseGPT}                              & 62.76 & 73.23 & 64.06 & 30.20 \\
\midrule
Magnitude                              & 17.50 & 70.08 & 58.04 & 25.68 \\
Random                                 & 31.21 & 70.89 & 59.55 & 28.07 \\
\texttt{Taylor}                                 & \textbf{59.23} & 72.91 & \textbf{63.85} & \textbf{29.69} \\
\texttt{GHOST}                                  & 51.76 & \textbf{73.12} & 63.72 & 29.10 \\
\bottomrule
\end{tabular}
\end{table}

\clearpage

\section{Extra Data.}
\label{app:extra_plots}

\begin{figure}[h]
\centering
\includegraphics[width=0.50\textwidth]{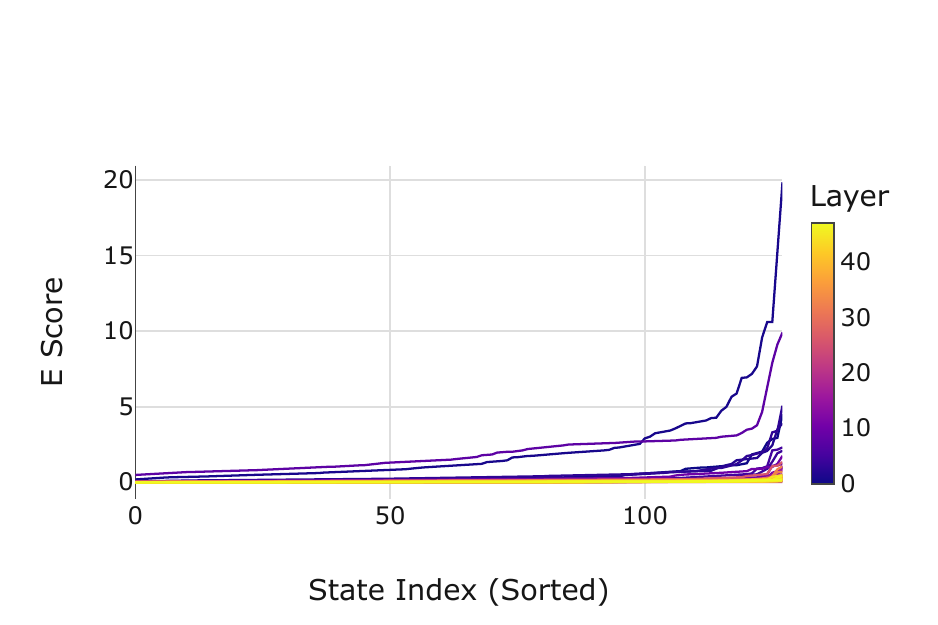}
\caption{\textbf{\texttt{GHOST} Energy Elbow.} An elbow plot of the ghost salience score on \texttt{Mamba2}-1.3B. As pictured, the elbow is around state index $110/128 \approx 86\%$ which is roughly when perplexity on the sparsity sweep, \Cref{tab:sparsity_sweep}, starts increasing too.}
\label{fig:energy_elbow}
\end{figure}

\begin{figure}[h]
\centering
\includegraphics[width=0.50\textwidth]{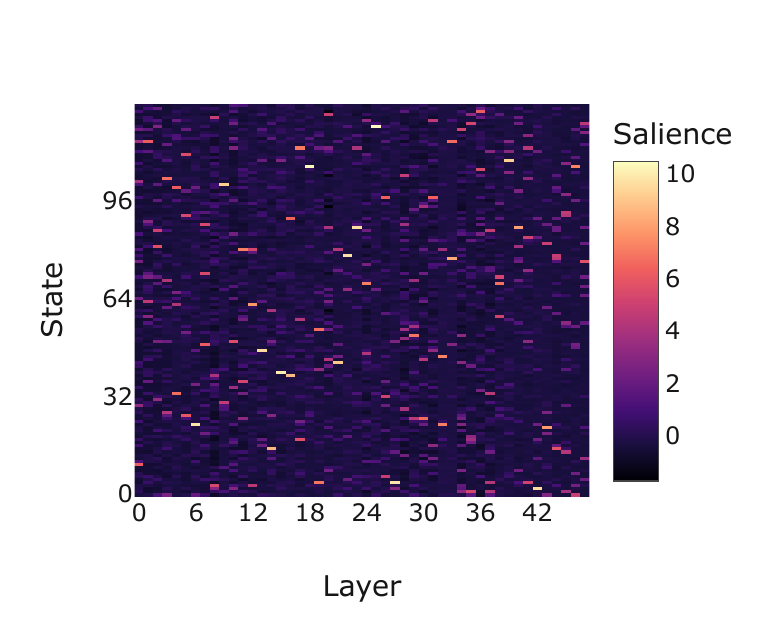}
\caption{\textbf{Energy-States Heatmap.} A heat map of \texttt{GHOST} salience score for each layer by state normalized by layer.}
\label{fig:energy_states_heatmap}
\end{figure}

\begin{table}[h]
\centering
\caption{\textbf{Architectural Robustness.} Evaluation of pruning methodologies on \texttt{Zamba2} hybrid architectures. We compare the perplexity of \texttt{Zamba2}-1.2B and \texttt{Zamba2}-2.7B across varying sparsity levels. In stark contrast to the pure \texttt{Mamba2} baseline in \Cref{tab:scaling}, \texttt{Taylor}-based pruning exhibits catastrophic failure on \texttt{Zamba2}, yielding random-guess perplexity ($>3\text{e}8$). We hypothesize this is driven by the interleaved shared attention blocks, which introduce unique structural vulnerabilities: 1) distortion of the gradient landscape which renders gradient-magnitude proxies like \texttt{Taylor} unreliable, and 2) the shared attention may inhibit adaptation to the activation distribution shifts caused by pruning the \texttt{Mamba2} backbone. Conversely, \texttt{GHOST} demonstrates resilience to this hybrid structure, maintaining perplexity near the dense baseline.}
\label{tab:zamba2_scaling}
\begin{tabular}{cccccc}
\toprule
{\textbf{Method}} & \textbf{1.2B}     & \textbf{2.7B}     \\
\midrule
Dense                                  & 11.15 & 19.79 \\
\texttt{SparseGPT}                              & 11.21 & 19.76 \\
\midrule
Magnitude                              & 1.0E5 & 51.89 \\
Random                                 & 451.7 & 51.46 \\
\texttt{Taylor}                                 & 3.0E8 & 3.5E5 \\
\texttt{GHOST}                         & \textbf{11.80}  & \textbf{21.23} \\
\bottomrule
\end{tabular}
\end{table}


\end{document}